\documentclass[onecolumn,10pt,aps,pre,amsmath,amsfonts,amssymb,floatfix,superscriptaddress,notitlepage,nofootinbib]{revtex4-2}
\usepackage[english]{babel}
\usepackage[utf8]{inputenc}
\usepackage{epsfig,amsmath,amssymb,amsfonts,graphicx,color,calc,epstopdf,bm}
\usepackage[T1]{fontenc}

\begin{document}
\title{Generative modeling through internal high-dimensional chaotic activity}
\author{Samantha J. Fournier}
    \email[Correspondence email address: ]{samantha.fournier@ipht.fr}
\author{Pierfrancesco Urbani}
    \affiliation{Université Paris Saclay, CNRS, CEA, Institut de physique théorique, F-91191 Gif-sur-Yvette, France}


\begin{abstract}
Generative modeling aims at producing new datapoints whose statistical properties resemble the ones in a training dataset. In recent years, there has been a burst of machine learning techniques and settings that can achieve this goal with remarkable performances.
In most of these settings, one uses the training dataset in conjunction with noise, which is added as a source of statistical variability and is essential for the generative task. 
Here, we explore the idea of using internal chaotic dynamics in high-dimensional chaotic systems as a way to generate new datapoints from a training dataset.
We show that simple learning rules can achieve this goal within a set of vanilla architectures and characterize the quality of the generated datapoints through standard accuracy measures.
\end{abstract}
\maketitle

\section{Introduction} Generative models aim at creating samples statistically similar to those belonging to a training dataset: their goal is to fit the probability distribution from which the datapoints supposedly come from. In a generic setting, this probability distribution takes the form of a Boltzmann factor. The corresponding Energy Based Models (EBMs) fit the parameters of the Hamiltonian of the Boltzmann distribution and can be viewed as  maximum entropy models, where the statistical properties of the dataset are imposed as constraints to low degree correlation functions \cite{jaynes_information_1957-1, jaynes_information_1957-2},   (see \cite{cocco2018inverse, mehta_high-bias_2019} for recent reviews). The resulting learning rule can be viewed as a gradient ascent on the Log-Likelihood (LL). However, running the training dynamics is a notoriously challenging task: at each {training step}, the evaluation of the gradient of the LL requires the computation of the correlation functions of the degrees of freedom as predicted from the current estimation of the model's probability distribution. This is typically an intractable problem from an analytical point of view and is generally {tackled numerically} through parallel Monte Carlo Markov Chain  (MCMC) simulations.
The main drawback of this strategy is that MCMC must be run for times that are larger than the mixing time, which can be very large especially in the low temperature phase and with datapoints clustered in high dimensions. Indeed in this case, the corresponding Boltzmann distributions are peaked on far-apart regions of phase space that are {separated by} free-energy barriers scaling with the dimension of the system. 
Exploring each of these regions requires activated dynamical trajectories, which implies that the MCMC is very slow to thermalize.
Therefore, training EBMs requires an extensive computational effort that–in practice–the corresponding MCMC is never thermalized and the gradient ascent on the LL is performed in an out-of-equilibrium fashion, with only a few Monte Carlo steps. 

A first approximate solution to this problem was proposed using clever initializations of the MCMCs \cite{hinton_article_2002, tieleman_training_2008}. However, recent works \cite{nijkamp_learning_2019, decelle_equilibrium_2022} have shown that  Restricted Boltzmann Machines (RBMs)–a particular class of EBMs–{trained with short-run, randomly initialized MCMCs can still generate good-looking samples}, provided that the schedule for the generation process coincides with the one used during the training dynamics. Therefore in these {(extreme)} experiments, the learning dynamics is not supposed to use a proxy for the exact gradient of the LL since the MCMC never gets to equilibrium. Correspondingly, the training trajectory is not maximizing the LL and yet the trained RBMs can generate good-looking samples.

\begin{figure*}[t]
    \centering
    \includegraphics[width=0.85\textwidth]{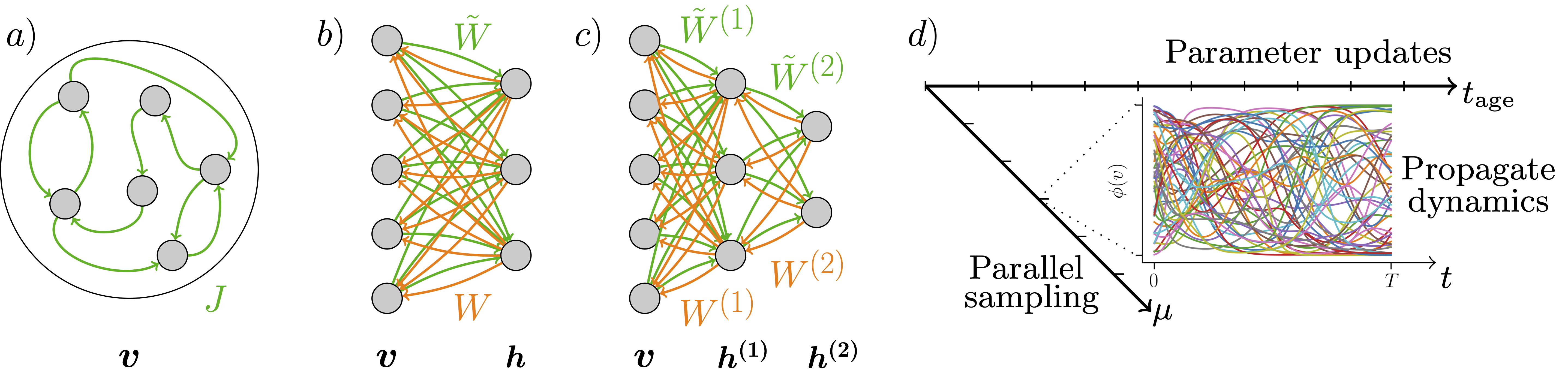}
    \caption{The three different architectures considered in this work: $a)$ Unrestricted architecture, $b)$ Restricted 2-layer architecture and $c)$ Restricted 3-layer architecture. The fields $\bm{b}$, $\bm{c}$ and $\bm{d}$ are not represented. Panel $d)$: Pipeline of training protocol.}
    \label{fig: architecture and pipeline}
\end{figure*}

These results can be reformulated by replacing the MCMC sampling dynamics with the {following} Langevin algorithm. Calling $-E$ the LL, one can replace MCMC with 
\begin{equation}
    \label{eq: MCMC Langevin dynamics}
    \begin{split}
    \frac{d\bm{x}}{dt} &= -\frac{\partial E}{\partial \bm{x}} + \bm{\eta}(t)
    \end{split}
\end{equation}
where $\bm \eta$ is just white noise and $\bm x$ are the dynamical degrees of freedom of which one would like to know specific correlation functions. 
The stationary probability distribution of eq.~\eqref{eq: MCMC Langevin dynamics} is the model's Boltzmann measure with energy $E$. However, this stationary measure is attained only asymptotically and the results of \cite{nijkamp_learning_2019, decelle_equilibrium_2022} showed that it is not necessary to wait for such long times: sampling the stochastic trajectories properly {(i.e. with random initializations)} on finite time windows and fitting according to the model's parameters, one can get an effective generative model. Therefore in this case, the system in \eqref{eq: MCMC Langevin dynamics} is held out-of-equilibrium in the sense that configurations are sampled before reaching the Boltzmann distribution. The question we are interested in in this work concerns how we can generalize the out-of-equilibrium dynamics and noise that can be used to generate new samples. This question is very broad and here we focus on the particular case in which the out-of-equilibrium dynamics is represented by a chaotic, high-dimensional activity induced by recurrent connections in the architecture of the generative models.
Our interest in this particular setting is related to neuroscience questions {\cite{amit_modeling_1989, dayan2005theoretical}}: we would like to try to understand if it is possible to train recurrent neural networks (RNNs) as generative models. In the following, we define a set of models and training algorithms with this purpose and present evidence that we can achieve via recurrent architectures good generative performances that we carefully quantify with standard accuracy indices.

\section{Definition of the models}
\label{sect: definition of the models}
In what follows, we consider the setting in which we have a training dataset  $\mathcal{D}=\{ \bm{\xi}^{(\mu)} \}_{\mu=1,\,...,\,N_s}$ composed by $N_s$ points which are $N_v$-dimensional vectors $\bm{\xi}^\mu$ and the generative task is to produce new datapoints that are statistically equivalent to the ones contained in the dataset.
The models we are going to define belong to a general class of dynamical systems which are high-dimensional and chaotic. 
We consider three architectures {of RNNs}, whose recurrent {and asymmetric} connections are used to {produce} chaotic dynamics which can be used for generative purposes. The parameters of the different models are trained with rules adapted from the usual contrastive Hebbian learning algorithms \cite{hinton_deterministic_1989}, where we replace the parallel MCMC sampling by parallel simulations of the dynamical system under study. The resulting generative models are straightforward to train and more biologically plausible than standard EBMs because they do not require the injection of noise by an external operator.

\textbf{Unrestricted architecture --} We consider one layer of $N_v$ variables $\bm{v} = \{v_i\}_{i=1,\dots,N_v}$ that are recurrently connected. They evolve according to
\begin{equation}
    \label{eq: dynamics of only visible and unrestricted model}
    \tau \frac{dv_i}{dt} = -v_i + \frac{1}{\sqrt{N_v}} \sum_{j=1}^{N_v} (J_{ij}+A_{ij}) \phi(v_j) + b_i
\end{equation}
where $\phi(.)=\mathrm{tanh(.)}$ is a non-linear activation function and $\tau$ fixes an intrinsic timescale in the dynamics. The  matrix $J$ is random with Gaussian independent entries with zero mean and variance $g$. We use $g$ as a control parameter: if $g$ is sufficiently large, the dynamical system with $A_{ij}=0$ and $b_i=0$ is in a chaotic phase \cite{sompolinsky1988chaos}. Therefore, the  matrix  $J$ will be used as a source of chaotic activity that drives the generative dynamics.   
The matrix $A$ and the field $\bm{b}$ are instead parameters that we train via contrastive {Hebbian learning}. 
They are both initialized to zero and at each training step $t_{\mathrm{age}}$ they are updated according to the rule
\begin{equation}
\label{eq: learning rules for unrestricted architecture}
\begin{split}
    \Delta A_{ij} &=k\left(\langle \phi(v_i) \phi(v_j) \rangle_\mathrm{clamped} - \langle \phi(v_i) \phi(v_j) \rangle_\mathrm{free}\right)\\
    \Delta b_i &=k\left( \langle \phi(v_i) \rangle_\mathrm{clamped} - \langle \phi(v_i) \rangle_\mathrm{free}\right),
\end{split}
\end{equation}
where $k$ is the learning rate.
$\langle . \rangle_\mathrm{clamped}$ is an empirical average over {a minibatch of size $M$ of} the training dataset $\{\bm{\xi}^{(\mu)} \}_{\mu=1,\,...,\,M}$, which is changed at each training step. $\langle . \rangle_\mathrm{free}$ is an empirical average over a set of $M$  configurations  $\{ \phi(\bm{v}^{(\mu)}(T)) \}_{\mu=1,\,...,\,M}$ collected by simulating the dynamical system \eqref{eq: dynamics of only visible and unrestricted model} up to time $T$, starting from independent random initial conditions. These two averages explicitly read for the update $\Delta A_{ij}$
\begin{gather*}
    \langle \phi(v_i) \phi(v_j) \rangle_\mathrm{clamped} = \frac1M \sum_{\mu=1}^M \xi_i^{(\mu)} \xi_j^{(\mu)}, \;\;\; \langle \phi(v_i) \phi(v_j) \rangle_\mathrm{free} = \frac1M \sum_{\mu=1}^M \phi(v_i^{(\mu)}(T)) \phi(v_j^{(\mu)}(T)).
\end{gather*}
Since the learning rule for $A$ is symmetric, training will suppress chaos by planting fixed points into the dynamics. The goal of the generative dynamics is therefore to find the optimal parameters $A$ and $\bm{b}$ that suppress chaos just enough to allow good generative performances. 

\begin{figure*}[t]
    \centering
    \includegraphics[width=\textwidth]{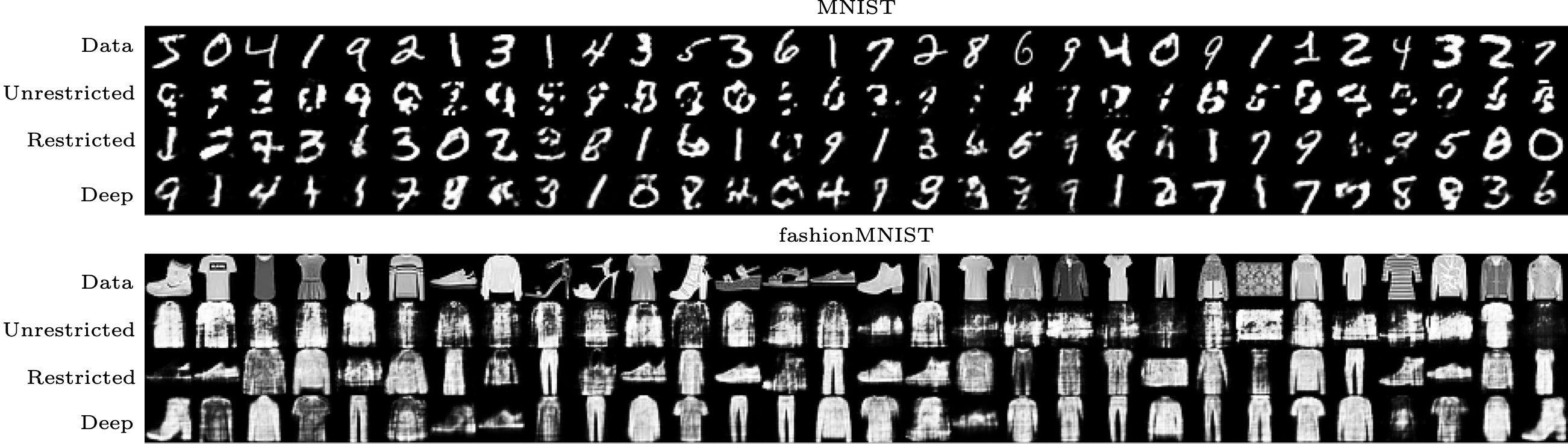}
    \caption{Data samples and generated samples from the trained models. Parameters: $N_v=784$ {(which is the dimension of each data sample)}, $N_h=N^{(1)}_h=500$, $N^{(2)}_h=100$, $dt=1$, $\tau=10$, $T=100$, $g=1.5$, $k=0.01$, $M=500$, $N_s=10\,000$. Training lasted $300\,000$ training steps in all cases except for the Deep model trained on FashionMNIST where training lasted $400\,000$ training steps.}
    \label{fig: MNIST and Fashion-MNIST generated samples}
\end{figure*}

\textbf{Restricted architecture --} In the context of EBMs, adding to the visible variables a set of latent variables allows to encode higher-order interactions in the former and this is a way of constructing more expressive generative models. We adapt the standard architecture of RBMs \cite{smolensky_information_1986} to construct a dynamical system composed of two layers of variables: the visible $\bm{v}$ and hidden $\bm{h}$ units, with no intra-layer connections. The variables are subjected to the dynamics
\begin{align}
    \label{eq : RBM visible units dynamics}
    \tau \frac{dv_i}{dt} &= -v_i + \frac{1}{\sqrt{N_h}} \sum_{a=1}^{N_h} \left( W_{ia} + A_{ia} \right) \phi(h_a) + b_i,\\
    \label{eq : RBM hidden units dynamics}
    \tau \frac{dh_a}{dt} &= -h_a + \frac{1}{\sqrt{N_v}} \sum_{i=1}^{N_v} \left( \tilde{W}_{ai} + A_{ai} \right) \phi(v_i) + c_a,
\end{align}
with $i=1,\, ...,\, N_v$ and $a=1,\, ...,\, N_h$. The matrix elements of the connections $W, \; \tilde{W}$ are drawn at random from two Gaussian distributions with mean $0$ and variance $g$. These matrices are fixed and the control parameter $g$ tunes the level of chaos in the untrained network. The fields $\bm{b},\, \bm{c}$ and the symmetric connections $A$ are parameters of the model that are initialized at zero and trained. The corresponding update rule, at each training step $t_{\mathrm{age}}$, is written as
\begin{equation}
\label{eq: learning rules for restricted architecture}
\begin{split}
    \Delta A_{ia} &=k\left( \langle \phi(v_i) \phi(h_a) \rangle_\mathrm{clamped} - \langle \phi(v_i) \phi(h_a) \rangle_\mathrm{free}\right)\\
    \Delta b_{i} &=k\left( \langle \phi(v_i) \rangle_\mathrm{clamped} - \langle \phi(v_i) \rangle_\mathrm{free}\right)\\
    \Delta c_a &=k\left( \langle \phi(h_a) \rangle_\mathrm{clamped} - \langle \phi(h_a)\rangle_\mathrm{free} \right),
\end{split}
\end{equation}
where $k$ is again the learning rate.
$\langle . \rangle_\mathrm{clamped}$ is now an empirical average over a set of $\mu=1,\ldots,M$ configurations collected from simulating the hidden units' dynamics \eqref{eq : RBM hidden units dynamics} up to time $T$, starting from independent random initial conditions, and while the visible units are clamped $\phi(\bm{v}^{(\mu)})(t) = \bm{\xi}^{(\mu)}$ to a data sample from a minibatch of the training dataset. In the particular case of a restricted architecture with no intra-layer connections, the \textit{clamped} dynamics doesn't need to be simulated since it can be integrated directly
\begin{equation}
    \label{eq: definition of chi variable}
	h_a^{(\mu, \mathrm{clamped})}(T) = h_a^{(\mu, \mathrm{clamped})}(0) e^{-T/\tau} + \left( \frac{1}{\sqrt{N_v}} \sum_{i=1}^{N_v} \left( \tilde{W}_{ai} + A_{ai} \right) \xi_i^{(\mu)} + c_a \right) \left( 1 - e^{-T/\tau} \right).
\end{equation}
On the other hand, $\langle . \rangle_\mathrm{free}$ is an empirical average over a set of $\mu=1,\,...,\,M$ configurations of the free dynamics $\{ \phi(\bm{v}^{(\mu)}(T)), \phi(\bm{h}^{(\mu)}(T)) \}_{\mu}$ collected by simulating the dynamical system \eqref{eq : RBM visible units dynamics}-\eqref{eq : RBM hidden units dynamics} up to time $T$, starting from independent random initial conditions. These two averages explicitly read for the update $\Delta A_{ia}$
\begin{gather*}
    \langle \phi(v_i) \phi(h_a) \rangle_\mathrm{clamped} = \frac1M \sum_{\mu=1}^M \xi_i^{(\mu)} \phi(h_a^{(\mu, \mathrm{clamped})}(T)), \;\;\; \langle \phi(v_i) \phi(v_j) \rangle_\mathrm{free} = \frac1M \sum_{\mu=1}^M \phi(v_i^{(\mu)}(T)) \phi(h_a^{(\mu)}(T)).
\end{gather*}
The crucial difference between the model considered here and standard RBM architectures is that here $\tilde{W} \neq W^T$ so the connections between layers are in general not symmetric. In the case of symmetric connections, one can see the dynamics in Eqs.~\eqref{eq : RBM visible units dynamics}-\eqref{eq : RBM hidden units dynamics} as some gradient descent on a carefully defined Lyapunov function \cite{krotov2020large}. In the asymmetric case instead, one expects chaotic activity \cite{blumenthal2024phase}, which can be used as a sort of generative noise.

\textbf{Deep and restricted architecture --} For the last architecture, we consider a generalization of the restricted one described above where we add an extra hidden layer to increase the depth of the model. This is introduced to capture more easily higher-order interactions between visible degrees of freedom. Therefore, we consider three layers of variables: $\bm{v}$, $\bm{h^{(1)}}$ and $\bm{h^{(2)}}$–still with no intra-layer connections. The precise architecture is represented in Fig. \ref{fig: architecture and pipeline}-c. The equations of motion are generalizations of Eqs.~\eqref{eq : RBM visible units dynamics}-\eqref{eq : RBM hidden units dynamics} {(see Appendix~\ref{appendix: deep architecture} for details)}. It must be noted that in this case the dynamics of the hidden neurons when the visible layer is clamped is not integrable and must be simulated as well.

\textbf{Training protocol --} The pipeline of training is summarized in Fig.~\ref{fig: architecture and pipeline}-d and is the same for all the architectures considered. At each training step $t_{\mathrm{age}}$, the parameters of the model considered need to be updated with the learning rules \eqref{eq: learning rules for unrestricted architecture} for the unrestricted architecture and \eqref{eq: learning rules for restricted architecture} for the restricted one. The positive part of the learning rule is computed on a minibatch of data $\{ \bm{\xi}^{(\mu)} \}_{\mu=1,\,...,\,M}$ of size $M$ that changes at every training step–without the need of simulating the \textit{clamped} dynamics of the system. Instead to compute the negative part of the learning rule, one needs to run $M$ parallel simulations of the \textit{free} dynamical system up to the target time $T$. To this end, the dynamical equations are Euler discretized with a timestep $dt$. The initial values of all degrees of freedom $\bm{x} = \{\bm{v}, \bm{h}\}$ are extracted at random from a normal probability distribution with mean $0$ and variance $1$. The parameters of the models are thus updated at each training step $t_{\mathrm{age}}$ with a learning rate $k$. This training strategy is similar to what was done in \cite{decelle_equilibrium_2022, nijkamp_learning_2019} where RBMs were trained using MCMCs initialized at random and simulated for oder $10$ to $100$ Monte Carlo steps. To gain efficiency, the dynamics is implemented with PyTorch and simulated on a GPU.

\begin{figure}[h]
    \centering
    \includegraphics[width=8cm]{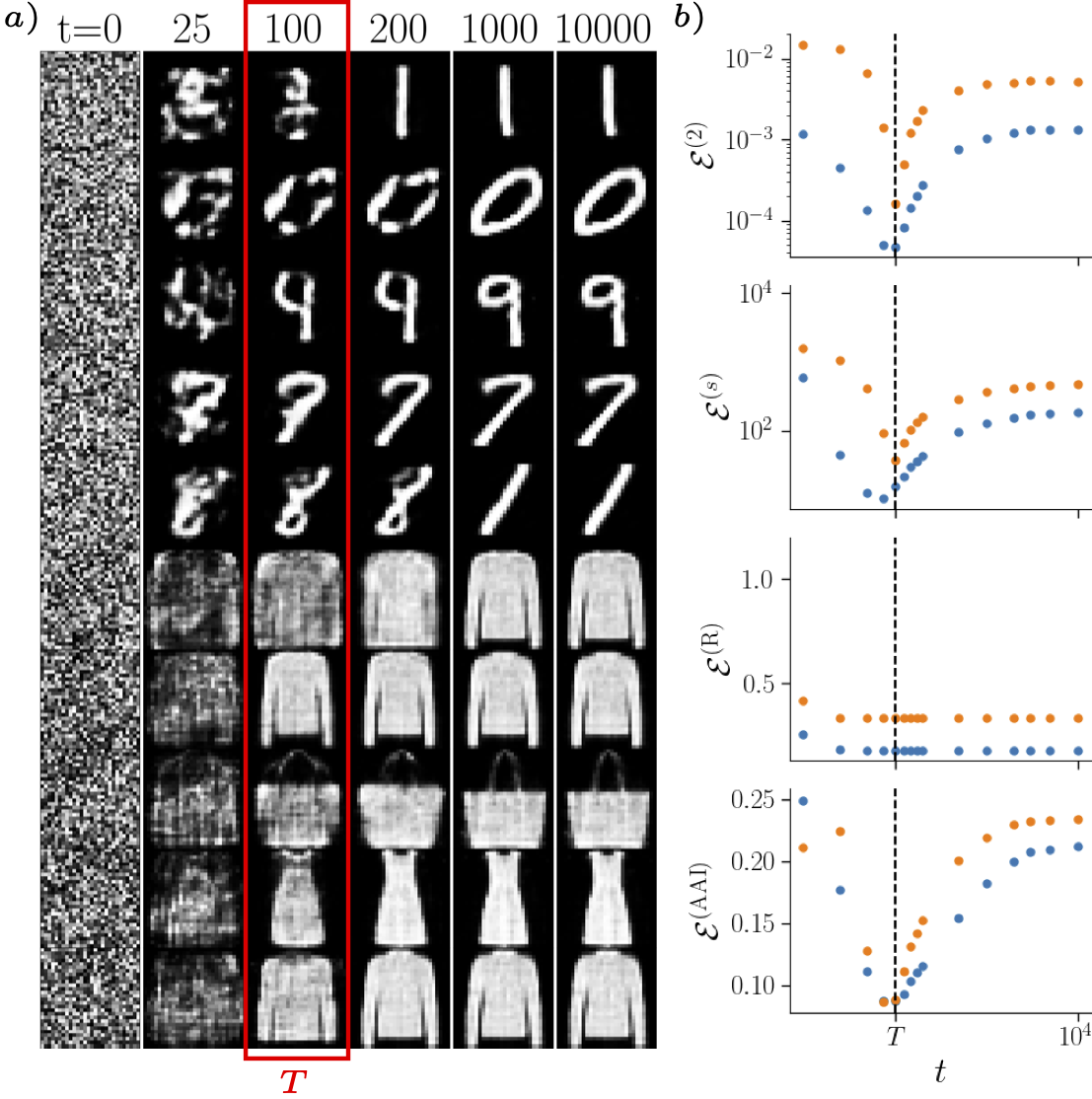}
    \caption{Panel $a)$ Samples generated at different times $t$ from the 2-layer Restricted model trained on MNIST (top half) and FashionMNIST (bottom half) with the same parameters as in Fig.~\ref{fig: MNIST and Fashion-MNIST generated samples}. Panel $b)$ Performance at different times $t$ of the latter model trained on MNIST (blue dots) and FashionMNIST (orange dots) across $4$ accuracy indices (see Appendix~\ref{appendix: accuracy indices} for their definitions). The lower these indices are, the better the generated samples. {All the accuracy indices are computed on a testing dataset {composed of $N_s=10\,000$ data samples}.}}
    \label{fig: Perfs at different times}
\end{figure}

\section{Results}
We tested the proposed models on the MNIST \cite{lecun1998mnist, lecun_gradient-based_1998} and Fashion-MNIST \cite{xiao_fashion-mnist_2017} datasets, which are composed of grey images $\bm{\xi} \in [0,\,255]^{N_v}$ of respectively handwritten digits and pieces of clothings. Since the dynamical systems are read trhough an activation function $\phi$ whose output is bounded between -1 and 1, we trained our models on transformed data samples $\bm{\xi} \to f(\bm{\xi}) = 2 \frac{\bm{\xi}}{255} - 1 \in [-1,\,1]^{N_v}$, and then performed the inverse transform to plot the generated samples as grey images. The samples generated by the trained models can be found in Fig.~\ref{fig: MNIST and Fashion-MNIST generated samples}: they correspond to the activity of the visible layer $\bm{v}$ at time $T$, starting from a random initialization. The 1-layer Unrestricted model generated samples that were recognizable but noisy overall. The 2- and 3-layers Restricted models generated samples that were qualitatively similar, and overall better than in the 1-layer case. While our motivation for the introduction of an additional layer to the Restricted model was to capture finer details of the data samples, we have seen that we did not get significant improvements in the quality of the generated samples. {Nevertheless, the 3-layer architecture performed sucessfully, given that it} was trained with vanilla learning rules and no special pre-training \cite{salakhutdinov_efficient_2010, hinton_better_2012}.

In Fig.~\ref{fig: Perfs at different times}-a, we show some snapshots of the dynamics of the trained 2-layer Restricted model at different times $t$, starting from a random initialization at $t=0$. During training, the dynamical system always learns a set of fixed-points that correspond to some ``good'' samples and that are reached for times $t>T$ (see for instance the samples generated at $t>1\,000$). These fixed-points {attractors} however are just a small fraction of the rich repertoire of samples that the trained models are able to generate at time $t=T$. In that respect, it is important to stress that the dynamics of the trained models at $t=T$ {has not yet reached a fixed-point}: the system is wandering in a subspace surrounding the fixed-points attractors. In Fig.~\ref{fig: Perfs at different times}-b, the statistical quality of the samples generated by the trained model is assessed through standard accuracy indices. The peak in performance at $t=T$ was also observed in \cite{decelle_equilibrium_2022} for RBMs trained with short-run, randomly initialized MCMCs. {We obtain similar values of $\mathcal{E}^{(2)}$ and $\mathcal{E}^{(\mathrm{AAI})}$ as Ref.~\cite{decelle_equilibrium_2022} at peak performance for a comparable number of training steps: see the \textit{dark blue} curves in Fig.~3 of Ref.~\cite{decelle_equilibrium_2022} Supplementary material, where the equivalent of $T$ is what the authors call $k$ (i.e. the number of Monte Carlo steps).}
Another interesting phenomenon already observed in \cite{decelle_equilibrium_2022} and also present in our setting is mode collapse: this can be clearly seen in the fifth row of Fig.~\ref{fig: Perfs at different times}-a {, where the generated sample at $t=T$ would have been classified as an $8$ but later evolves to become a $1$}. We empirically observe that this also happens with models trained on FashionMNIST, but less frequently.

\begin{figure}[h]
    \centering
    \includegraphics[width=6cm]{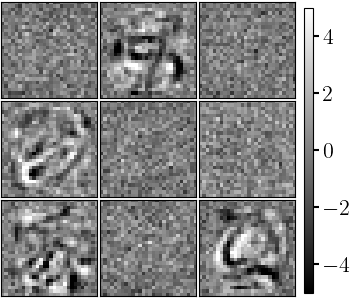}
    \caption{Receptive fields of $9$ randomly chosen hidden neurons in the 2-layer Restricted version trained on MNIST. {Note that the overall color scale has been divided by $g$.}}
    \label{fig: receptive fields}
\end{figure}
Lastly in Fig.~\ref{fig: receptive fields}, we show the receptive fields of hidden neurons of the 2-layer Restricted model trained on MNIST. Specifically, we plot for some indices $a \in [1, \, N_h]$ the weights $\{ W_{ia} + A_{ia} \}_{i=1,\,...,\,N_v}$ connecting the hidden unit $h_a$ to the visible layer $\bm{v}$. After training, about half of the hidden neurons have spatially structured receptive fields, as is typically the case in trained RBMs \cite{montavon_practical_2012, hinton_article_2002}, while the other half have random receptive fields not much different than their initial state when $A_{ai}=0$. {After running the dynamics of the trained networks with these hidden units clamped to 0, it seems that they encode for the edges of the MNIST images which are always black.}

\section{Conclusions}
In this work, we investigated the possibility of using high-dimensional chaotic activity to train recurrent dynamical systems as generative models. The resulting models are autonomous, i.e. they do not require an external noise injection. Furthermore, the learning rules are essentially Hebbian, i.e. they depend on the correlation between pre- and post-connecting neurons. Thus the proposed generative models could be an interesting starting point towards more biologically plausible generative models.

Our implementation uses vanilla learning rules that are most likely sub-optimal. It would be interesting to see if modifying the learning rules with standard tricks like regularization, momentum or centering of gradients \cite{montavon_practical_2012, melchior_how_2016} would improve the quality of the generated samples. Another interesting perspective would be to modify the learning rule by applying a non-linear function to the Hebbian terms, as was done in Dense Associative Memory networks \cite{krotov2020large}, to see if this would improve the memory capacity and thus the expressivity of the generative model.

Finally, the study presented here heavily relied on numerical simulations. A theoretical approach to the questions discussed in this manuscript is possible by combining simplified high-dimensional non-linear chaotic systems and dynamical mean field theory \cite{fournier2023statistical}.

{\bf Acknowledgments -- } This work was supported by the French government under the France 2030 program (PhOM - Graduate School of Physics) under reference ANR-11-IDEX-0003.

\bibliographystyle{unsrt}
\bibliography{library.bib}

\appendix
\section{Definition of the deep restricted architecture}
\label{appendix: deep architecture}
The dynamical equations of the 3-layer Restricted architecture read
\begin{align}
    \label{eq : Deep v dynamics}
    &\tau \frac{dv_i}{dt} = -v_i + \frac{1}{\sqrt{N^{(1)}_h}} \sum_{a=1}^{N^{(1)}_h} \left( W^{(1)}_{ia} + A^{(1)}_{ia} \right) \phi(h^{(1)}_a) + b_i,\\
    \label{eq : Deep h1 dynamics}
        &\tau \frac{dh^{(1)}_a}{dt} = -h^{(1)}_a + \frac{1}{\sqrt{N_v}} \sum_{i=1}^{N_v} \left( \tilde{W}^{(1)}_{ai} + A^{(1)}_{ai} \right) \phi(v_i) + \frac{1}{\sqrt{N^{(2)}_h}} \sum_{b=1}^{N^{(2)}_h} \left( W^{(2)}_{ab} + A^{(2)}_{ab} \right) \phi(h^{(2)}_b) + c_a,\\
    \label{eq : Deep h2 dynamics}
    &\tau \frac{dh^{(2)}_a}{dt} = -h^{(2)}_a + \frac{1}{\sqrt{N^{(1)}_h}} \sum_{b=1}^{N^{(1)}_h} \left( \tilde{W}^{(2)}_{ab} + A^{(2)}_{ab} \right) \phi(h^{(1)}_b) + d_a,
\end{align}
where the matrix elements of the connections $W^{(1)},\,W^{(2)},\,\tilde{W}^{(1)},\,\tilde{W}^{(2)}$ are drawn at random from four Gaussian distributions with mean $0$ and variance $g$. The parameters $A^{(1)}$, $A^{(2)}$, $\bm{b}$, $\bm{c}$ and $\bm{d}$ are initialized at zero and updated at each training step $t_{\mathrm{age}}$ with
\begin{align}
    \Delta A^{(1)}_{ia} &= k \left( \langle \phi(v_i) \phi(h^{(1)}_a) \rangle_\mathrm{clamped} - \langle \phi(v_i) \phi(h^{(1)}_a) \rangle_\mathrm{free} \right)\\
    \Delta A^{(2)}_{ab} &= k \left( \langle \phi(h^{(1)}_a) \phi(h^{(2)}_b) \rangle_\mathrm{clamped} - \langle \phi(h^{(1)}_a) \phi(h^{(2)}_b) \rangle_\mathrm{free} \right)\\
    \Delta b_{i} &= k \left( \langle \phi(v_i) \rangle_\mathrm{clamped} - \langle \phi(v_i) \rangle_\mathrm{free} \right)\\
    \Delta c_a &= k \left( \langle \phi(h^{(1)}_a) \rangle_\mathrm{clamped} - \langle \phi(h^{(1)}_a) \rangle_\mathrm{free} \right)\\
    \Delta d_a &= k \left( \langle \phi(h^{(2)}_a) \rangle_\mathrm{clamped} - \langle \phi(h^{(2)}_a) \rangle_\mathrm{free} \right),
\end{align}
where $i=1,\, ...,\, N_v$, $a=1,\, ...,\, N^{(1)}_h$ and $b=1,\, ...,\, N^{(2)}_h$. As for the unrestricted and 2-layer restricted architectures, $\langle . \rangle_\mathrm{clamped}$ is an empirical average over $\mu=1,\dots,M$ configurations obtained by simulating the dynamics \eqref{eq : Deep h1 dynamics}-\eqref{eq : Deep h2 dynamics} up to the target time $T$, starting from different initial conditions, and while the visible layer is clamped to a data sample $\phi(\bm{v}^{(\mu)})(t) = \bm{\xi}^{(\mu)}$ from a minibatch of the training dataset. Instead $\langle . \rangle_\mathrm{free}$ is an empirical average over $\mu=1,\dots,M$ configurations obtained by simulating the \textit{free} dynamics \eqref{eq : Deep h1 dynamics}-\eqref{eq : Deep h2 dynamics} up to the target time $T$, starting from different initial conditions. $k$ is the learning rate.

\section{Definition of the accuracy indices}
\label{appendix: accuracy indices}

\begin{figure}
    \centering
    \includegraphics[width=0.9\textwidth]{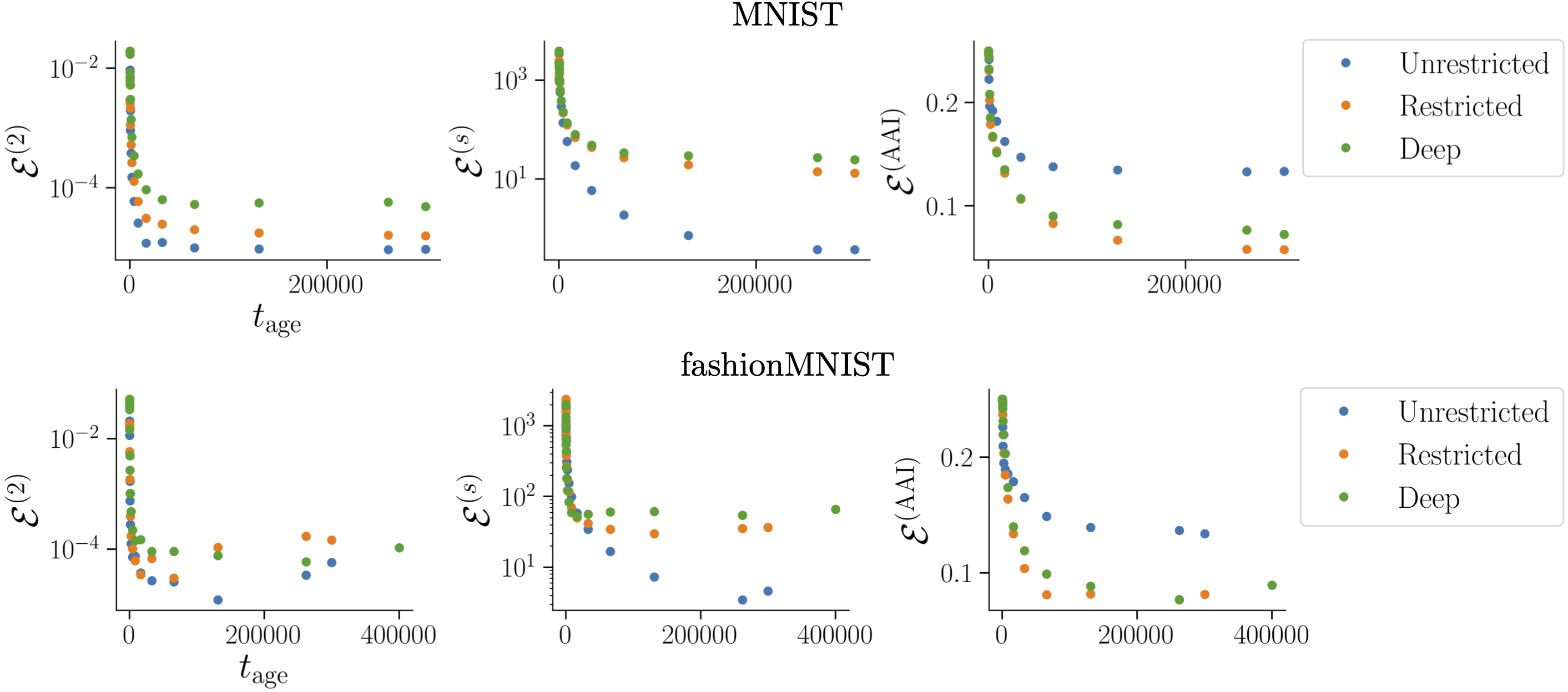}
    \caption{Performance of the different models trained on MNIST (top half) and FashionMNIST (bottom half) during training ($t_\mathrm{age}$ is the number of training steps). {The accuracy indicators are computed at the target time $T^*=T$.}}
    \label{fig: Perfs during learning}
\end{figure}
Monitoring the performance of a generative model is not a straightforward task because it is unclear what would be a precise definition of a good generated sample in the case in which the probability distribution of the underlying dataset is unknown.
To assess the statistical quality of generated samples, we therefore considered four possible quality measures that were previously studied in the literature \cite{decelle_equilibrium_2022, carbone_fast_2023}.

The first one is the error on the second moment $\mathcal{E}^{(2)}$ between the generated and the original datapoints. We define the averaged covariance matrix of the visible units $C_{ij} = \overline{\mathrm{Cov}(\phi(v_i),\phi(v_j))}$ (where $\overline{(...)}$ is an average of $N_s$ independent samples). We compute this quantity both on samples from the dataset $\mathcal{D}$ setting $\phi(\bm{v})=\bm{\xi}$, and on generated samples $\mathcal{G}$. Here $\mathcal{G}=\{ \phi(\bm{v}^{(\mu)})(T^*) \}_{\mu=1,\,...,\,N_s}$ is obtained by running $N_s$ parallel simulations of the free dynamical system up to a target time $T^*$, starting from a random initialization. We then define $\mathcal{E}^{(2)}$ as the Mean Squared Error (MSE) of their difference:
\begin{equation}
    \mathcal{E}^{(2)} = \frac1{N_v(N_v-1)} \sum_{i<j} \left( C_{ij}^{\mathcal{G}} - C_{ij}^{\mathcal{D}} \right)^2.
\end{equation}
The smaller $\mathcal{E}^{(2)}$, the better the statistical distribution of the generated samples.

Next, we define the error on the spectrum $\mathcal{E}^{(s)}$. Let's consider a matrix of samples $X \in \mathbb{R}^{N_s \times N_v}$. Its singular value decomposition (SVD) can be written as $X = U S V^T$, where $U \in \mathbb{R}^{N_s \times N_s}$, $V \in \mathbb{R}^{N_v \times N_v}$ and $S = \mathrm{diag}(\{s_{\mu}\})$, with the singular values $s_{\mu}$ ordered such that $s_1>s_2>...>s_{N_s}$. These singular values can be computed on a matrix of data samples or of generated samples, so we define $\mathcal{E}^{(s)}$ as
\begin{equation}
    \mathcal{E}^{(s)} = \frac1{N_s} \sum_{\mu=1}^{N_s} \left( s_{\mu}^{\mathcal{G}} - s_{\mu}^{\mathcal{D}} \right)^2.
\end{equation}

For the 2-layer Restricted version of the model, it is also common to look at the so-called reconstruction error $\mathcal{E}^{(R)}$. The idea is to see how well the model can reconstruct a given data sample. To do this, the 2-layer Restricted model is presented with a data sample $\bm{\xi}^{(\mu)}$ (for $\mu \in \{1,\,...,\,N_s\}$), from which we compute the activities of the hidden units at time $T^*$, while the visible units are clamped to the data sample $\phi(\bm{v}^{(\mu)}) = \bm{\xi}^{(\mu)}$. Next, we do the inverse operation: we fix the hidden units to their previously calculated value $\bm{h}^{(\mu)}(T^*)$ and predict the activity of the visible units at time $T^*$. The latter is the generated prediction of the model $\phi(\bm{v}^{(\mu)})(T^*)$, which is to be compared with the original data sample $\bm{\xi}^{(\mu)}$. The reconstruction error is then defined as the MSE of their difference, averaged over visible sites $i$ and independent samples $\mu$:
\begin{equation}
    \mathcal{E}^{(R)} = \frac1{N_s N_v} \sum_{\mu=1}^{N_s} \sum_{i=1}^{N_v} \left( \xi^{(\mu)}_i - \phi(v^{(\mu)}_i) \right)^2.
\end{equation}

Lastly, we define the error on the Adversarial Accuracy Indicator (AAI). This score was originally introduced in \cite{yale_generation_2020}. The goal with this indicator is to measure how well data samples and generated samples are mixed with one another. We start by creating a matrix of samples $X \in \mathbb{R}^{2N_s \times N_v}$ by stacking together the data and the generated samples. We then create a matrix $D$ whose entries are the euclidean distance between each pair of samples in $X$. By taking the minimum of each rows of $D$, we can compute 
$P_{\mathcal{G} \mathcal{G}}$ and $P_{\mathcal{D} \mathcal{D}}$ defined respectively as the probability that a generated sample has a nearest neighbor (n.n.) which is also a generated sample, and the probability that a data sample has a n.n. which is also a data sample. If the generated and data samples are perfectly mixed, $P_{\mathcal{G} \mathcal{G}}$ and $P_{\mathcal{D} \mathcal{D}}$ should both be $0.5$, so we define $\mathcal{E}^{(AAI)}$ as:
\begin{equation}
    \mathcal{E}^{(AAI)} = 0.5 \left[ (P_{\mathcal{G} \mathcal{G}}-0.5)^2 + (P_{\mathcal{D} \mathcal{D}}-0.5)^2 \right].
\end{equation}

These quality measures are computed over a batch of samples generated from the dynamical system at a target time $T^*$, which can be equal to $T$ or generic. The behaviour of these accuracy indicators computed from samples generated by a trained model as a function of a generic $T^*=t$ is reported Fig.~\ref{fig: Perfs at different times}-b. Instead, the behaviour of these accuracy indicators computed from samples generated at $T^*=T$ as a function of the training step $t_{\mathrm{age}}$ are shown Fig.~\ref{fig: Perfs during learning} ($\mathcal{E}^{(R)}$ is not shown). We observe that all the error functions decrease steeply at the beginning of training and then saturate to small values. It is interesting to note that the error functions are not always in agreement with regards to which model architecture performs best. These error functions are therefore just a coarse way of quantifying the performance of a generative model.

\end{document}